\documentclass[runningheads]{llncs}
\usepackage{graphicx}

\usepackage{lineno}
\usepackage{tikz}
\usepackage{comment}
\usepackage{amsmath,amssymb} 
\usepackage{color}
\usepackage{hyperref}
\usepackage{multirow}
\usepackage[accsupp]{axessibility}  

\usepackage[width=122mm,left=12mm,paperwidth=146mm,height=193mm,top=12mm,paperheight=217mm]{geometry}
\usepackage{ulem}
\usepackage{subfigure}

\newcommand{\hide}[1]
     {
     }

\begin{document}
\nolinenumbers
\pagestyle{headings}
\mainmatter
\def\ECCVSubNumber{100}  

\title{RepBNN: towards a precise Binary Neural Network with Enhanced Feature Map via Repeating} 

\titlerunning{ECCV-22 submission ID \ECCVSubNumber} 
\authorrunning{ECCV-22 submission ID \ECCVSubNumber} 
\author{}
\institute{Paper ID \ECCVSubNumber}

\titlerunning{Abbreviated paper title}
%
\author{Xulong Shi\inst{1} \and
Zhi Qi\inst{1} \and
Jiaxuan Cai \inst{1}\and
Keqi Fu \inst{1}\and
Yaru Zhao \inst{1}\and
Zan Li \inst{1}\and
Xuanyu Liu \inst{1}\and
Hao Liu \inst{1}}
\authorrunning{F. Author et al.}
%
\institute{National ASIC System Engineering Research Center, Southeast University, China
\email{\{long,101011256\}@seu.edu.cn}
}

\maketitle

\begin{abstract}
 Binary neural network (BNN) is an extreme quantization version of convolutional neural networks (CNNs) with all features and weights mapped to just 1-bit. Although BNN saves a lot of memory and computation demand to make CNN applicable on edge or mobile devices, BNN suffers the drop of network performance due to the reduced representation capability after binarization. In this paper, we propose a new replaceable and easy-to-use convolution module RepConv, which enhances feature maps through replicating input or output along channel dimension by $\beta$ times without extra cost on the number of parameters and convolutional computation. We also define a set of RepTran rules to use RepConv throughout BNN modules like binary convolution, fully connected layer and batch normalization. Experiments demonstrate that after the RepTran transformation, a set of highly cited BNNs have achieved universally better performance than the original BNN versions. For example, the Top-1 accuracy of Rep-ReCU-ResNet-20, \textit{i.e.}, a RepBconv enhanced ReCU-ResNet-20\cite{xu2021recu}, reaches 88.97\% on CIFAR-10, which is 1.47\% higher than that of the original network. And Rep-AdamBNN-ReActNet-A \cite{liu2020reactnet}\cite{liu2021adambnn} achieves 71.342\% Top-1 accuracy on ImageNet, a fresh state-of-the-art result of BNNs. Code and models are available at: \href{https://github.com/imfinethanks/Rep_AdamBNN}{https://github.com/imfinethanks/Rep\_AdamBNN}.

\hide{Recently, Binary Neural Networks(BNNs) have made it possible to deploy DNN in embedded devices by replacing expensive full-precision MACs with XNOR and bit-count. However, it also leads to a significant decrease in the accuracy, due to the reduced representation capability. The most direct solution is to increase the number of channels. This paper proposes a new convolutional structure, named RepConv, increasing the number of channels without extra convolution OPs and parameters. On this basis, a set of universal network structure transformation rules are proposed, called RepTran, which can transform other popular BNNs into the style of RepBNN (BNN with RepConv) easily. We apply RepTran to a large number of state-of-the-art binarization works, leading to a series of enhanced binary networks, which are validated on CIFAR-10\cite{krizhevsky2009cifar10} and ImageNet\cite{russakovsky2015imagenet}. Among them Rep-AdamBNN-ReActNet-A(RepTran enhanced AdamBNN-ReActNet-A\cite{liu2020reactnet}\cite{liu2021adambnn}) achieves 71.34\% Top-1 precision on ImageNet without extra cost on convolution OPs. 
}
\end{abstract}

\section{Introduction}\label{section:Intro}
As we all know, the applications of convolutional neural networks (CNNs) have achieved tremendous success in computer vision fields such as image classification, object detection, object tracking, and depth estimation. But this success comes at the cost of a huge amount of computation, which binds the application of convolutional neural networks generally to high-end hardware, such as GPU, TPU, etc. However, the efficient application of CNN models on mobile devices and embedded devices, where the storage space and computing resource are limited, is still quite challenging. In order to solve this problem, various lightweight network technologies have emerged, mainly including network architecture design\cite{sandler2018mobilenetv2}\cite{zhang2018shufflenet}, network architecture search\cite{howard2019mobilenetv3}\cite{liu2018darts}, knowledge distillation\cite{hinton2015distilling}, pruning\cite{liu2017pruning}\cite{ding2019pruning}, and quantization\cite{zhou2016dorefa}\cite{yang2020searching}.

\hide{The method that quantizes high-precision CNN weights and feature maps to low-precision is called quantization. }Among these lightweight network technologies, binarization contributes to an extreme version of a convolutional neural network\hide{The most extreme quantization method is binarization}, where all feature maps and weights are represented by just 1-bit. Binarization greatly improves the efficiency of CNN models, because it not only reduces the volume of parameters, the requirement of storage capacity, but also its replacement of MAC operations with efficient XNOR and bit-count operations saves a lot computational expense. The work in \cite{rastegari2016xnor} shows that a binary neural network has achieve 32× parameter compression and 58× speedup than its full-precision network.

However, the speedup of calculation and the compression of parameter by binarization have obvious impact on the accuracy of CNN models. It is the declined representation capability of feature maps that degrades the performance of binarized CNNs. Unlike full-precision networks, the value range of binarized convolutions is so extremely restricted \cite{liu2018birealnet} that the accuracy loss becomes more serious. If the number of input channels is $C_{in}$, and the size of the convolution kernel is $k_{h}$ and $k_{w}$, then the value range of the binary convolution is $\left (-C_{in}*k_{h}*k_{w}, C_{in}*k_{h}*k_{w}\right)$, a total of $C_{in}*k_{h}*k_{w} +1$ quantization levels. Obviously, increasing the number of input channels $C_{in}$ can improve the feature representation capability of binary neurual networks (BNNs).

Inspired by this observation, we propose a new replaceable convolution module RepConv, which reshapes the weight kernel through expanding its channel dimension while compressing its number of groups. The total number of parameters and the amount of related convolutional calculation remain unchanged after the reshape.  To convolve with such a reshaped weight kernel, RepConv has to increase the number of both input and output channels simply through a replication operation. In such a manner, RepConv enriches the network information. 

The binary version of RepConv is called RepBconv. If replacing the normal binarized convolution with RepBconv and further modify the remaining modules such as the first convolution layer, batch normalization, bypasses and fully connected layer following the transformation rule of RepTran, we translate a set of highly cited binary neural networks~\cite{xu2021recu}\cite{liu2020reactnet}\cite{liu2021adambnn} into the corresponding RepBNNs, which have been demonstrated to achieve universally better performance than the original BNN versions.

\begin{figure}[h]
\begin{minipage}[h]{.6\linewidth}
\centering
\includegraphics[width=6cm]{ 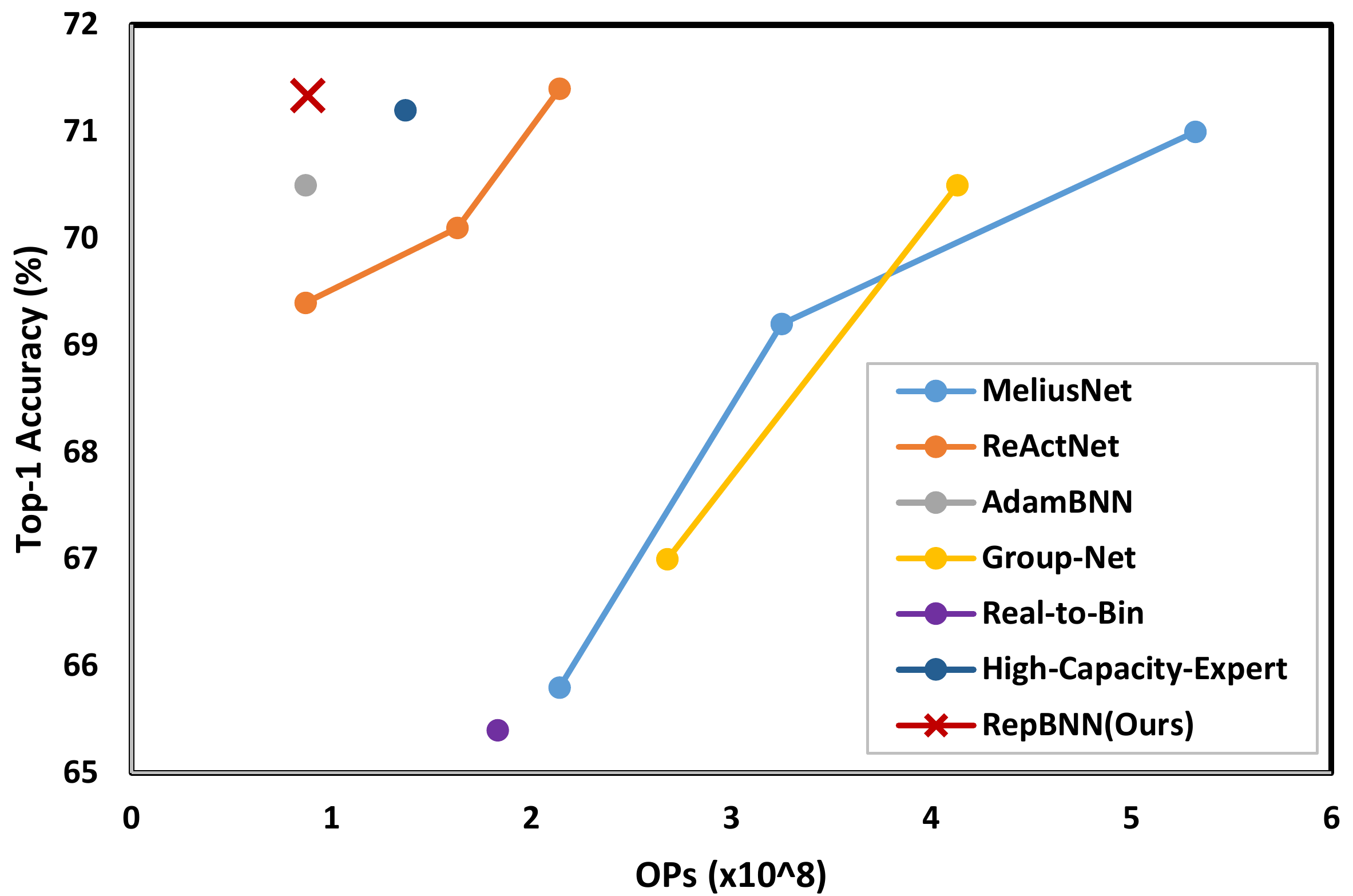}
\end{minipage}%
\begin{minipage}[h]{.4\linewidth}\tiny
\centering
\begin{tabular}{l|l|l}
\hline
Methods                    & OPs(x$10^{8}$) & Top-1(\%) \\ \hline
Real-to-Bin\cite{martinez2020real-to-binary}                & 1.83           & 65.4          \\ \hline
\multirow{2}{*}{Group-Net\cite{zhuang2019Group-Net}} & 2.68           & 67.0          \\ \cline{2-3} 
                           & 4.13           & 70.5          \\ \hline
\multirow{3}{*}{MeliusNet\cite{bethge2020meliusnet}} & 2.14           & 65.8          \\ \cline{2-3} 
                           & 3.25           & 69.2          \\ \cline{2-3} 
                           & 5.32           & 71.0          \\ \hline
\multirow{3}{*}{ReActNet\cite{liu2020reactnet}}  & 0.87           & 69.4          \\ \cline{2-3} 
                           & 1.63           & 70.1          \\ \cline{2-3} 
                           & 2.14           & 71.4          \\ \hline
AdamBNN\cite{liu2021adambnn}                    & 0.87           & 70.5          \\ \hline
High-capacity\cite{bulat2020high}       & 1.37           & 71.2          \\ \hline
\textbf{RepBNN(Ours)}      & \textbf{0.88}  & \textbf{71.3} \\ \hline
\end{tabular}

\end{minipage}
\caption{OPs vs. ImageNet Top-1 Accuracy. The OPs of some methods is referenced from \cite{yuan2021review}. RepBNN achieves state-of-the-art result with 71.34\% top-1 accuracy on ImageNet. With similar accuracy, RepBNN is 2.4x more efficient than ReActNet. With similar OPs, RepBNN is 0.84\% more accurate than AdamBNN.    }
\label{fig:Fig.1}
\end{figure}

Experimental results show that after the RepTran transformation, 
the Top-1 accuracy of Rep-ReCU-ResNet-20, \textit{i.e.}, a RepBconv enhanced ReCU-ResNet-20\cite{xu2021recu}, reaches 88.97\% on CIFAR-10, which is 1.47\% higher than that of the original network. And Rep-AdamBNN-ReActNet-A \cite{liu2020reactnet}\cite{liu2021adambnn} achieves 71.342\% 
Top-1 accuracy on ImageNet, a fresh state-of-the-art result of BNNs as shown in Fig.\ref{fig:Fig.1}. 

To sum up, this paper makes the following contributions:

\begin{itemize}
    \item[•] We propose a new and easy-to-use convolution module named RepConv that reshapes the weight kernels thus increases the number of input and output channels so as to enrich the representation capability of binarized feature map without extra cost on convolutional operations and parameters.
    \item[•] We propose a set of universal network transformation rules of RepTran, which makes the\hide{other general }common modules in binary neural network adaptable to the application of RepConv. 
    \item[•] We apply RepTran to a large number of state-of-the-art binary neural networks and validate them on CIFAR-10 and ImageNet extensively. It is proved that our proposed structure can significantly improve the performance of binary networks.
\end{itemize}

\section{Related Work}

Binarization greatly reduces the amount of operations (OPs) and parameters, but encounters a severe drop in accuracy, which is mainly caused by the loss of information accompanied in the network with just 1-bit weights and activations. Previous work mitigated this precision decline through either an efficient training procedure that accurately back-propagates gradient values or methods to improve the representation capability of binary neural networks.

\subsection{Training}
The impact of loss to weights is hard to be propagated backward through the partial derivative of the sign function that has an infinite gradient value at zero but a constant elsewhere\hide{ is hard to propagate the impact of loss gradient backwards to weights}. Thus the first BNN work~\cite{courbariaux2016bnns} estimated gradients of STE\cite{bengio2013STE} instead of the sign function during back-propagation. Bi-Real Net\cite{liu2018birealnet} replaced STE with a piecewise linear function, which is a second-order approximation of the sign function. IR-Net\cite{qin2020irnet} employed Error Decay Estimator (EDE) to minimize information loss during the back propagation, thus ensured adequate updates at the beginning of training. RBNN\cite{lin2020rbnn} proposed a training-aware approximation of the sign function \hide{which used in}for gradient backpropagation. Real-to-Bin\cite{martinez2020real-to-binary}designed a sequence of teacher-student pairs to bridge the architectural gap between real and binary networks. Lately ReActNet\cite{liu2020reactnet} adopted a distribution loss that measures the distribution similarity between the binary and real-valued network. ReCU\cite{xu2021recu} explored the effect of dead weights and introduced a rectified clamp unit (ReCU) to revive it. AdamBNN\cite{liu2021adambnn} expounded the influence of Adam and weight decay when training BNNs and proposed a better training strategy. These recent work promoted the performance of BNNs to a significantly higher level than the first BNN~\cite{courbariaux2016bnns}.

\subsection{Representation Capability}

Improving the representation capability of models is another major research direction of binary networks. XNOR-Net\cite{rastegari2016xnor} used two channel-wise scaling factor for activations and weights to estimate the real-value. The scaling factor here is often replaced by batch normalization in practical implementation and this technique is widely used by subsequent work. Another popular work named Bi-Real Net\cite{liu2018birealnet} added residuals to each layer to make the full-precision data flow compensating the information loss in the binarized main branch of the network\hide{, which is also widely used in subsequent work}. When designing the transformation rule of RepTran, we have considered the modules of both batch normalization and residuals, so that our work is applicable to a wide range of available binary neural networks.\hide{Our method takes the advantage of both batch normalization and residuals, and the broad use of both greatly provides the generality of our work, making our work applicable to almost any binary neural network. }Real-to-Bin\cite{martinez2020real-to-binary} used a data-driven channel re-scaling gated residual leading to a superior performance. Group-Net\cite{zhuang2019Group-Net} and BENN\cite{zhu2019benn} combined multiple models to trade the amount of computation for accuracy. ReActNet\cite{liu2020reactnet}, based on an improved mobilenet model, used RPRelu and RSign to explicitly shift and reshape the activation distribution, further reducing the performance gap between a BNN and its full-precision network.

Applying our technique of RepBconv following the transformation rule of RepTran to these great previous work will raise the accuracy to a newly high level.
\hide{The method in this paper starts from improving the representation capability of the model, and can be used in combination with other works to achieve state-of-the-art performance.}

\section{The Module of RepConv}
\label{section:repconv}
As explained in Section~\ref{section:Intro}, increasing the number of input channels can improve the feature representation capability of BNNs. Inspired by this observation, we propose a new convolution module named RepConv.

\begin{figure}[htbp]
    \centering
    \includegraphics[width=1.0\textwidth]{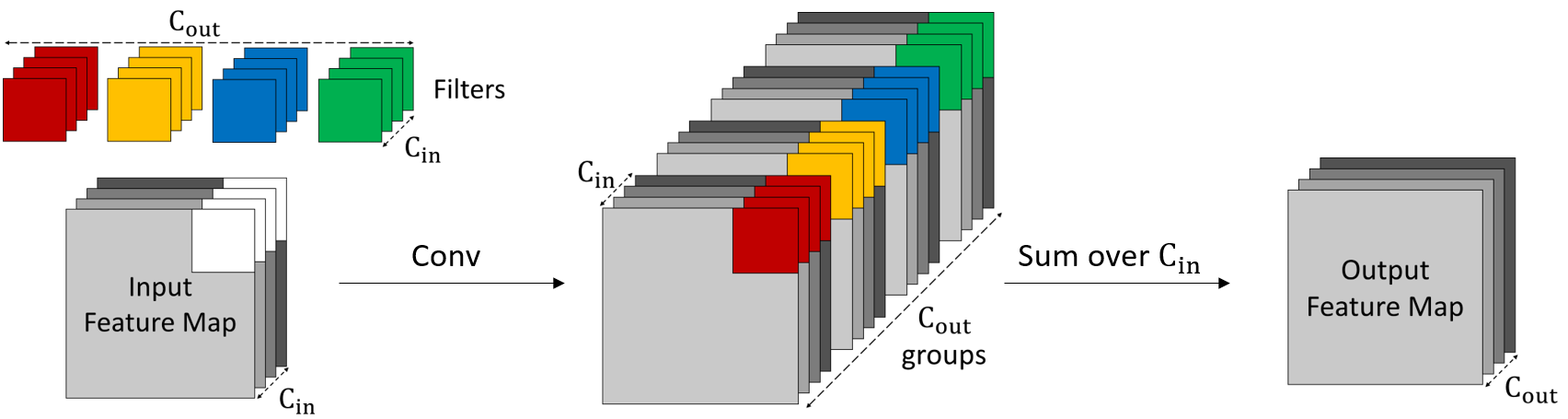}
    \caption{A normal convolution process}
    \label{fig:Fig.2}
\end{figure}

Fig.\ref{fig:Fig.2} shows a normal convolution process. In Fig.\ref{fig:Fig.2}, the structure of a convolution kernel is expressed as $\left ( C_{out}, C_{in},k_{h},k_{w} \right )$ , where the number of input channels is $C_{in}$, and the number of output channels is $C_{out}$.

\begin{figure}[htbp]
    \centering
    \includegraphics[width=1.0\textwidth]{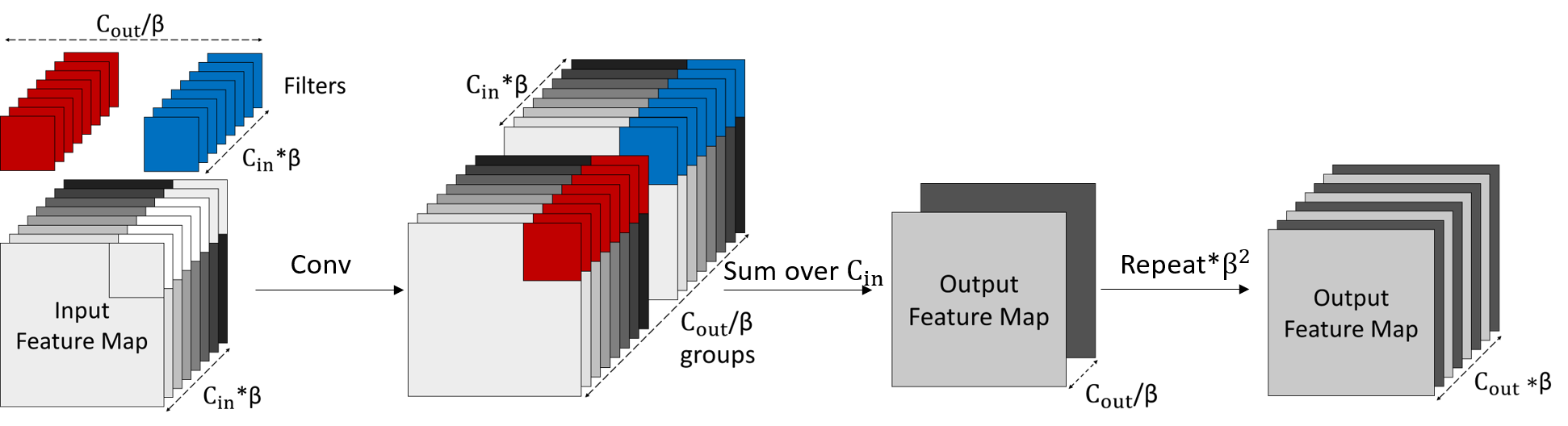}
    \caption{The convolution process of RepConv}
    \label{fig:Fig.3}
\end{figure}

RepConv reshapes the convolution kernel to $\left ( C_{out}/\beta, C_{in}*\beta,k_{h},k_{w} \right )$ with $C_{in}*\beta$ input channels and $C_{out}/\beta$ output channels respectively. Then RepConv makes $\beta^{2}$ copies of the output feature map and concates them along the channel direction, so that the final channelwise size of the output becomes $C_{out}*\beta$. This step is called a repeat operation. The specific process of RepConv can be seen in Fig.\ref{fig:Fig.3}.

Finally, at the same cost of convolutional computation, RepConv generates feature map of $C_{out}*\beta$ channels,  while its counterpart, a normal convolution process, creates a feature map of just $C_{out}$ channels. Since the output of a previous layer is used as the input to the next layer, the expanded information is consistently passed on in networks using RepConv. A batch normalization right after the repeat operation of RepConv and a residual link that transfers the output of last layer to that of this layer make full use of the enriched feature map by RepConv.

Our experiments have demonstrated that the replacement of normal convolution module with RepConv improves the accuracy for both full-precision and binary networks.  More discuss on why a simple repeating operation improves the network accuracy and how the modules of batch normalization and residual link impact the feature contents after RepConv is presented in Section \ref{subsection:bn_and_res}. 

\hide{channels is increased by a factor of $\beta$ with the same budget of weights and at the same cost of convolutional computation as in normal convolution process.}

$\beta$ is a hyperparameter of RepConv that represents the degree of information dilation in a feature map. However a larger number of $\beta$ raises up the computational overhead of non-convolutional operations like batch normalization. In Section \ref{section:computational}, we give an example to demonstrate this computational overhead of RepConv is negligible to the overall calculation of the entire network. The discuss on how different $\beta$  affects the accuracy is given in Session~\ref{subsection:Configuration}.

\section{RepTran}
\begin{figure}[htbp]
    \centering
    \subfigure[ A normal binarized convolution structure with regular Bconv module]{
    \includegraphics[width=9cm]{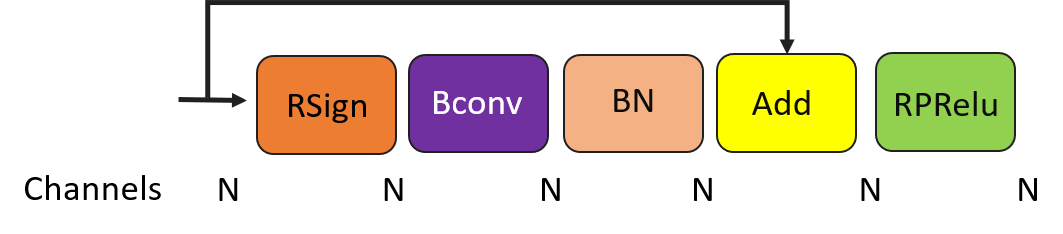}
    }
    \subfigure[A RepBconv module]{
    \includegraphics[width=4.5cm]{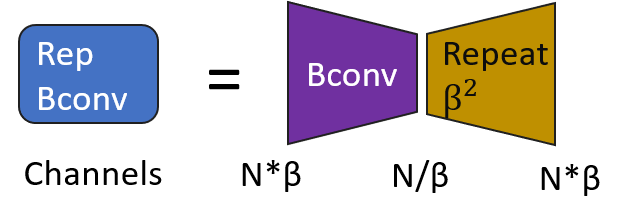}
    }
    \subfigure[A binary convolution structure after the replacement of BConv with RepBconv]{
    \includegraphics[width=9cm]{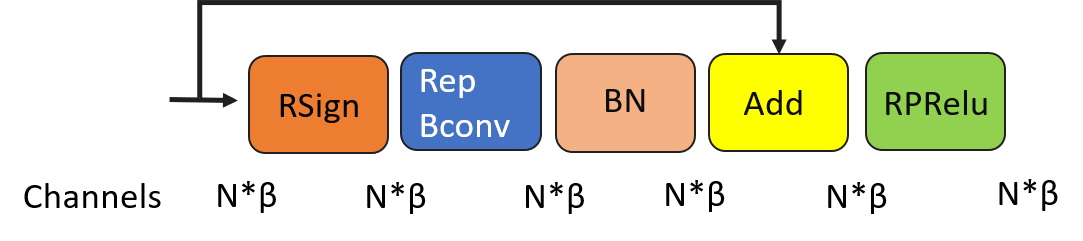}
    }
    \caption{Taking a structure fragment in ReActNet as an example, the usage of RepBconv module brings more feature information to traditional binary neural networks.}
    \label{fig:Fig.4}
\end{figure}

\hide{The binary version of RepConv is called RepBconv, as shown in Fig.\ref{fig:Fig.4}b. We exchange the regular binarized convolution module with RepBconv, then the size of the feature map through all the passes in the binary convolution structure is enlarged by $\beta$ times as in Fig.\ref{fig:Fig.4}c. } 

The binary version of RepConv is called RepBconv in Fig.\ref{fig:Fig.4}b, which consists a regular Bconv module with reshaped weight kernels and a subsequent repeating $\beta^2$x operation. A complete BNN generally consists structures such as the first full-precision convolution layer, binary convolution layers, bypasses, batch normalizations, and the last fully connected layer. 
As shown in Fig.\ref{fig:Fig.4}c, the usage of RepBconv instead of the regular BConv increases the size of information flow by $\beta$ times in a binary convolution structure and so as to the entire BNN network. 

\begin{figure}[htbp]
    \centering
    \includegraphics[width=1.0\textwidth]{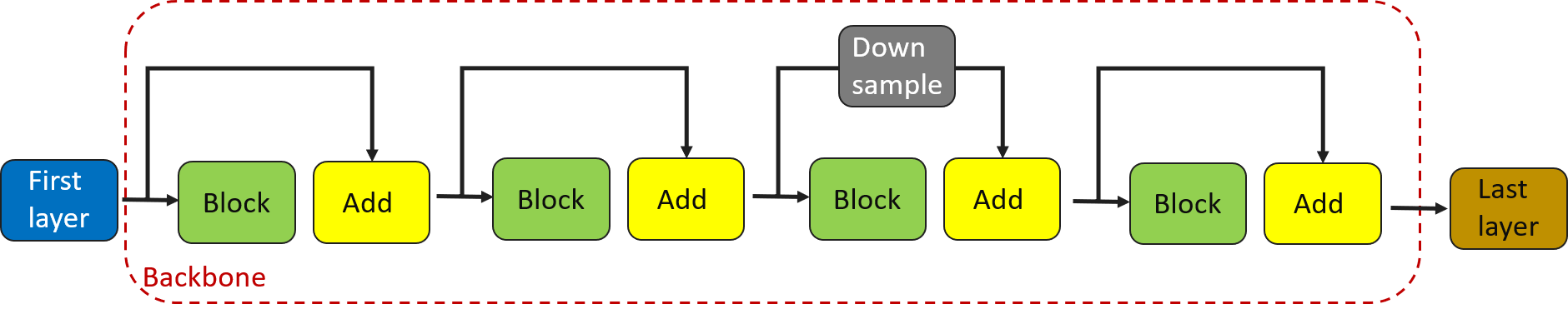}
    \caption{ A diagram of a typical binary neural network consisting a first layer, a last layer, and a backbone in between.}
    \label{fig:Fig.5}
\end{figure}

The transformation rule for these BNN structures accompanying with the usage of RepBconv is called RepTran. When the number of input and output channels is increased by RepBconv, these structures must be changed accordingly to maintain the function while control the extra computation cost related to the enlarged information size. This section explain how RepTran works from the first layer, through the backbone and until the last layer of the network, as illustrated in Fig.~\ref{fig:Fig.5}.

\hide{Except binary convolution, a complete BNN generally has structures such as the first layer of full-precision convolution, bypass, batch normalization, and the last fully connected layer. When the number of input and output channels is increased by RepBconv, these structures must be changed accordingly as well. The transformation rule for these BNN structures accompanying with the usage of RepBconv is called RepTran. This section explain how RepTran works from the first layer, through the backbone and until the last layer of the network, as illustrated in Fig.~\ref{fig:Fig.5}.}

\subsection{The Rule of RepTran in the First Layer}

Most of the BNNs use floating-point weights and activations in first layers to avoid the loss of network accuracy, such as ReCU\cite{xu2021recu}, ReActNet\cite{liu2020reactnet}, and RBNN\cite{lin2020rbnn}. There are also works like FracBNN\cite{zhang2021fracbnn} that encodes the information in the first layer and binarizes the rest part of the network.

No matter how these first layers are implemented, RepTran just replicates $\beta$ times of its output feature map along the channel direction right before features enter the batch normalization module. The rule of RepTran in the first layer is to prepare a $\beta$ fold block of input for the implementation of RepBconv in successive layers without interrupting the original structure of the first layer.

\subsection{The Rule of RepTran in the Backbone}
As shown in Fig.\ref{fig:Fig.5}, a backbone is the main body of a typical binary neural network that consumes the most of the computation. A backbone consists of a series of binarized convolution structures with necessary residual bypasses. Some bypasses just deliver data to Add modules, while some other bypass may down-sample the data necessarily. The computational expense of the downsampling bypass that uses full-precision convolution depends on the size of downsampling inputs and outputs.

In the main branch of a backbone, RepTran simply replaces all the Bconv modules with RepBconvs.  Details of this replacement is explained in Section~\ref{section:repconv}. However, RepTran transforms the downsampling bypass using full-precision convolution differently. \hide{Since RepTran increases the overall number of channels of the binary network by a factor of $\beta$, the computation amount of full-precision convolution in downsampling  will be $\beta^{2}$ fold. Therefore, }In order to avoid the $\beta^2$ times convolutional calculation due to the enlarged size of inputs and outputs by RepBconv, RepTran has to replace the regular full-precision convolution in a downsampling bypass with full-precision RepConv. For other operations in backbone, such as Sign, batch normalization and Relu, RepTran increases the number of input channels by a factor of $\beta$ directly.

\begin{figure}[htbp]
    \centering
    \includegraphics[width=1\textwidth]{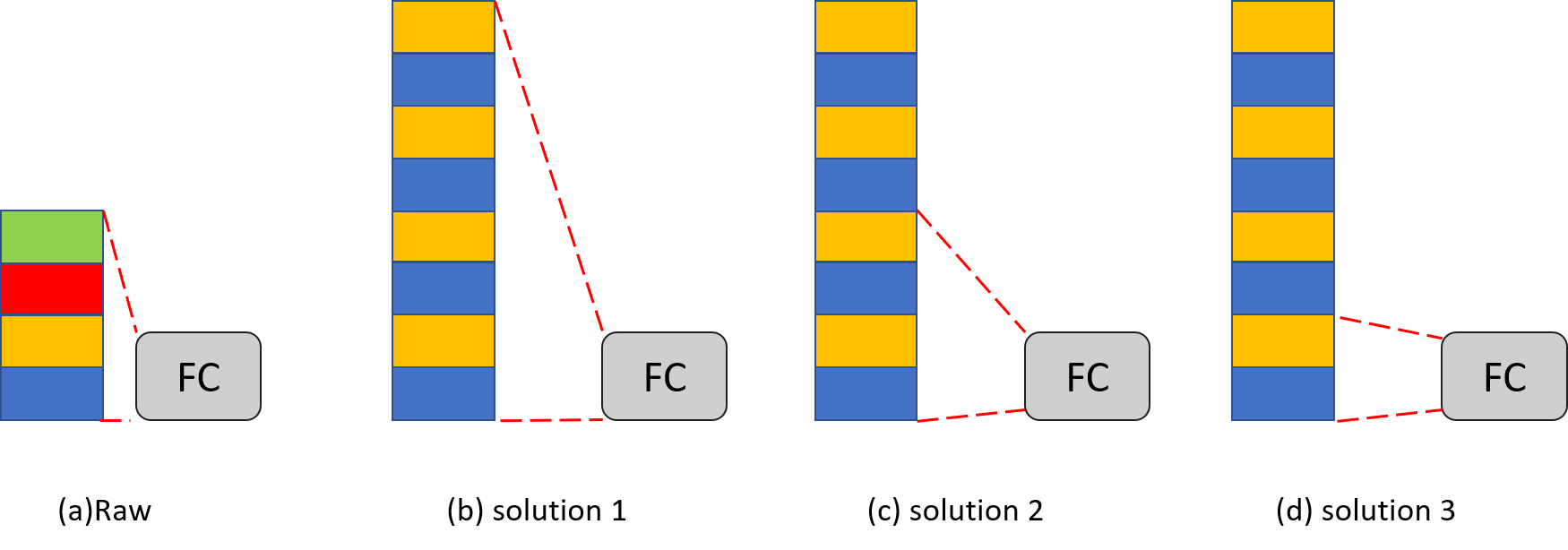}
    \caption{Three different ways of RepTran in the last layer with $\beta$=2. a) the original last layer, b) RepTran takes all the input channels, c) RepTran takes $1/\beta$ of the input channels, d)  RepTran takes $1/\beta^2$ of the input channels.}
    \label{fig:Fig.6}
\end{figure}

\subsection{The Rule of RepTran in the Last Layer}
\label{subsection:LastLayer}

Most of the BNNs take a full-precision fully connected layer as the last layer, which consumes full-precision operations. Given the input with $\beta$ times channels than that in the original network, a fully connected layer consequently consumes $\beta$ fold more parameters and computation.

RepTran in the last layer may be implemented in three ways. First,it uses all the channels of the input that increases the amount of computation by $\beta$ times. Second, it uses $1/\beta$ channels of the input and fulfills the computation at the same budget as the original final layer. Third, it uses $1/\beta^2$ channels of the input that reduces computation amount. Without losing generality, we exhibit the three different ways to run RepTran with $\beta=2$ in the last layer in Fig.\ref{fig:Fig.6}. According to the experimental results in Section \ref{section:ablation}, the RepTran rule transforms the last layer by the 1st way, which achieves better accuracy at the cost of a small amount of increment in computation.

\subsection{The Extra Computational Expense of RepTran }
\label{section:computational}
RepTran does not consume more convolution calculation than the original network. However, the expanded size of the information flow after applying RepConv causes extra computation in modules like Sign, Relu, and more computationally expensive fully connected layer (FC) or Batch Normalization (BN). \hide{Since BN and FC are much more computationally expensive than Sign and Relu, we only analyze the increased calculation amount in BN and FC due to RepTran.}

The BN operation can be divided into two steps as normalization and scaling, each with equal amount of calculation in a normal binarized convolution structure. Since RepConv or RepBconv enrich the number of output channels through replication, the normalization result based on just one set of the channels is shared by the others. Therefore the computation consumed by normalization is reduced to 1/2$\beta$ of the original.  However, the calculation amount of scaling, which is proportional to the number of channels, is increased by $\beta$ times. In summary, after the RepTran, the total amount of BN calculation becomes 1/2$\beta$+$\beta$/2 of the original.

The statistics of OPs in ReactNet-A\cite{liu2020reactnet} before and after RepTran ($\beta$=2) is listed in Table.\ref{tab:Tab.1}. Following the methods in~\cite{rastegari2016xnor}\cite{liu2020reactnet}\cite{liu2018birealnet}, we count the total number of operations as $OPs = FLOPs+BOPs/64$, where $FLOPs$ and $BOPs$ mean the  computation amount for floating-point and binary respectively. \hide{Since the previous work\cite{liu2020reactnet} does not count the OPs of BN, we separately list OPs with and without BN. }In the case of RepTran of $\beta=2$, the calculation of FC has been doubled and that of BN has been raised up by 25\%. However the total OPs of the entire ReactNet-A is 29x than the extra computational cost of FC and BN after RepTran. The analysis based on the network of ReactNet-A in Table.\ref{tab:Tab.1} demonstrates the computation overhead when implementing RepBconv through RepTran is negligible to the overall calculation of the entire network.

\begin{table}[]
\caption{Statistics of OPs of ReActNet-A before and after RepTran ($\beta$=2). FC: fully connected layer. Conv: full-precision convolution. BN: batch normalization. Bconv: binarized convolution. Among them, FC, Conv, and BN are floating-point operations, and Bconv is a binary operation. $OPs = FLOPs+BOPs/64$}
\begin{tabular}{l|lll|l|ll}
\hline
\multirow{2}{*}{} & \multicolumn{3}{l|}{FLOPs(x$10^{7}$)}                           & BOPs(x$10^{9}$) & \multicolumn{2}{l}{OPs(x$10^{8}$)}                \\ \cline{2-7} 
                  & \multicolumn{1}{l|}{FC}    & \multicolumn{1}{l|}{Conv}  & BN    & Bconv           & \multicolumn{1}{l|}{OPs-without-BN} & OPs-with-BN \\ \hline
ReActNet-A        & \multicolumn{1}{l|}{0.102} & \multicolumn{1}{l|}{1.084} & 1.009 & 4.822           & \multicolumn{1}{l|}{0.872}          & 0.973       \\
Rep-ReActNet-A    & \multicolumn{1}{l|}{0.205} & \multicolumn{1}{l|}{1.084} & 1.261 & 4.822           & \multicolumn{1}{l|}{0.882}          & 1.008       \\
$\Delta$          & \multicolumn{1}{l|}{0.102} & \multicolumn{1}{l|}{-}     & 0.252 & {-}               & \multicolumn{1}{l|}{0.010}          & 0.035       \\ \hline
\end{tabular}
\label{tab:Tab.1}
\end{table}

\section{Experiments}

\hide{In this section, we will first discuss the reasons why RepBNN achieves high accuracy in section \ref{subsection:bn_and_res}. Section \ref{subsection:Configuration} will also discuss the impact of different configurations of RepTran. Then we will discuss the performance of RepConv in full-precision network in section \ref{section:full-precision}. Finally to verify the advantages of RepBNN, section \ref{section:comparison} will apply RepTran on a large number of state-of-the-art binarization works and compare their accuracies on CIFAR-10 and ImageNet. }

For a fair comparison, we use the open source inferencing and training codes provided by IR-Net\cite{qin2020irnet}, RBNN\cite{lin2020rbnn}, FracBNN\cite{zhang2021fracbnn}, ReCU\cite{xu2021recu}, Bi-Real Net\cite{liu2018birealnet}, ReActNet\cite{liu2020reactnet} and AdamBNN\cite{liu2021adambnn}. We also apply exactly the same settings in data preprocessing as these recent outstanding BNN works.

\subsection{Ablation Studies}
\label{section:ablation}

\subsubsection{Configuration of RepTran in Fully Connected Layer}
\label{subsection:Configuration}  should be discussed.

There are three different ways to apply RepTran in a fully connected layers as introduced in section \ref{subsection:LastLayer}. As illustrated by the benchmark on top of basic BNN work as IRNet-ResNet-20 and ReCU-ResNet-20 in Table.\ref{tab:Tab.4}, the first solution of RepTran provides the best accuracy on CIFAR-10 at about 1k extra cost of OPs, which is acceptable. 

\begin{table}[]
\caption{Benchmark three different RepTran solutions in a fully connected layer on top of IRNet-ResNet-20 and ReCU-ResNet-20@ CIFAR-10 in terms of Top-1$\%$ and OPs. }
\centering
\begin{tabular}{l|ll|ll|ll|ll}
\hline
\multirow{2}{*}{} & \multicolumn{2}{l|}{Raw} & \multicolumn{2}{l|}{solution 1} & \multicolumn{2}{l|}{solution 2} & \multicolumn{2}{l}{solution 3} \\ \cline{2-9} 
                  & Top-1(\%)         & OPs        & Top-1(\%)                & OPs             & Top-1(\%)             & OPs          & Top-1(\%)                & OPs              \\ \hline
IR-Net            & 86.50        & 1069696
            & 87.59              &  1070336
               & 87.50            & 1069696
             & 87.50               &  1069376
                \\
ReCU              & 87.50        &  1069696          & 88.97                   &     1070336            & 88.92           &    1069696          & 88.82                   &        1069376         \\ \hline
\end{tabular}
\label{tab:Tab.4}
\end{table}

\subsubsection{The choice of Hyperparameter $\beta$ }
\label{subsection:beta value} balances the computational overhead of non-convolutional operations, the storage requirement for duplicated information, and the accuracy of RepBNNs. In the ablation study of $\beta$, each layer of the network shares the same $\beta$ value. Table.\ref{tab:Tab.5} shows the effect of $\beta$s on the accuracy of IRNet-ResNet-20 and ReCU-ResNet-20@ CIFAR-10. Although $\beta=4$ achieves a better accuracy, we set $\beta$ to 2 by default in RepBconv for the best trade-off between the additional OPs, the limited storage capacity and the gain of accuracy. 
 
\hide{ a limited OPs by BN and FC modules
 it also introduces more OPs as discussed in Section \ref{section:computational}. For a fair comparison, RepTran sets $\beta$ to 2 by default.}

\begin{table}[]
\caption{The accuracy of ResNet-20 @ CIFAR-10 with different $\beta$s on top of IRNet and ReCU.}
\centering
\begin{tabular}{l|l|l|l|l}
\hline
       & Raw(\%)  & $\beta$=2(\%) & $\beta$=4(\%) & $\beta$=8(\%) \\ \hline
IR-Net & 86.50 & 87.59      & 88.07     & 87.07     \\
ReCU   & 87.50 & 88.97     & 89.59     & 88.72    \\ \hline
\end{tabular}
\label{tab:Tab.5}
\end{table}

\hide{\subsubsection{The Importance of the Number of Input Channels to a Binarized Convolution.}

As stated in Section~\ref{section:Intro}, increasing the number of input channels increases quantization levels in the output feature map of the binary network, thereby improving accuracy. To demonstrate this we compare the performance of IRNet and ReCU on cifar10 and ResNet-20 after applying RepTran and group convolution respectively. The $\beta$ of RepTran is set to 2, and the group convolution doubles the number of network channels and sets the number of groups to 4. Compared with the original network, they have the same amount of convolution computation, the main difference is that the number of input channels of repbnn is twice that of the original network, and the group convolution is 1/2. Table \ref{tab:Tab.9} shows the experimental results, RepTran improves the accuracy of the network, while group convolution brings a huge drop. In the case of the same amount of convolution calculation, the more input channels, the higher the accuracy of the binary network.}

\hide{\begin{table}[]
\caption{The accuracy of ResNet-20 @ CIFAR-10 of IRNet and ReCU applying RepTran, group convolution respectively }
\centering
\begin{tabular}{l|l|l|l}
\hline
       & Raw(\%) & RepBNN(\%) & Group Convolution(\%) \\ \hline
IR-Net & 86.50   & 87.59      & 81.28                 \\ 
ReCU   & 87.50   & 88.97      & 83.12                 \\ \hline         
\end{tabular}
\label{tab:Tab.9}
\end{table}}

\subsubsection{The Role of Batch Normalization and Residuals to RepBconv}
\label{subsection:bn_and_res}should be explored. First, we discuss the position to insert BN module to RepBconv. As shown in Fig.\ref{fig:Fig.4}b, we can insert a BN before or after the repeat operation. Table.\ref{tab:Tab.2} provides the experimental results on IRNet-ResNet-20 and ReCU-ResNet-20. Applying the batch normalization before repeat operation downgrades the accuracy by $2\sim3\%$, however the accuracy is improved by about 1\% if running BN on the extended outputs through repeating.   

\hide{and their RepBNN version ($\beta$=2) with batch normalization before or after repeating the data. The best accuracy by BN after repeating the data reveals its contribution to the performance improvements of RepBNNs. }

\begin{table}[]
\caption{The accuracy of batch normalization before and after repeat in Rep-IRNet-ResNet-20 ($\beta$=2) and Rep-ReCU-ResNet-20 ($\beta$=2) on CIFAR-10 }
\centering
\begin{tabular}{l|l|l|l}
\hline
       & original(\%)  & RepBNN(BN after repeat)(\%) & RepBNN(BN before repeat)(\%) \\ \hline
IR-Net & 86.50 & 87.59   & 82.74                    \\
ReCU   & 87.50 & 88.97  & 85.30                  \\ \hline
\end{tabular}
\label{tab:Tab.2}
\end{table}

In order to verify the role of residual link as the bypass shown in Fig.\ref{fig:Fig.4}c to RepBcov,  we benchmarks the accuracy on CIFAR-10 for RepTran transformed IR-Net and ReCU on top of ResNet-20 and VGG respectively. The experimental results in Table.\ref{tab:Tab.3} show that RepTran stably improves the accuracy of Rep-IRNet-ResNet-20 and Rep-ReCU-ResNet-20, but has little effect or even decreases the accuracy on Rep-IRNet-VGG and Rep-ReCU-VGG, which is free of residual links.

\begin{table}[]
\caption{The accuracy of VGG and ResNet-20 before and after RepTran on CIFAR-10}
\centering
\begin{tabular}{l|l|l|l}
\hline
                  & Raw(\%)   & RepBNN(\%) & $\Delta(\%)$                     \\ \hline
IR-Net-VGG        & 90.40  & 86.95  & {\color[HTML]{9A0000} -3.45} \\
ReCU-VGG          & 92.22 & 92.41  & {\color[HTML]{009901} 0.19}  \\ \hline
IR-Net-ResNet-20   & 86.50  & 87.59   & {\color[HTML]{009901} 1.09}     \\
ReCU-ResNet-20 & 87.50  & 88.97  & {\color[HTML]{009901} 1.47}  \\ \hline
\end{tabular}
\label{tab:Tab.3}
\end{table}

The experimental results in Table.\ref{tab:Tab.2} and Table.\ref{tab:Tab.3} inspire us that if we transform a BNN with residual links to RepBNN and apply the batch normalization after the repeat operation of RepBconv,  the accuracy of the network will be improved. 

\hide{An obvious advantage of RepConv is that the number of channels in the network is increased by a factor of $\beta$. But simply repeating the same data doesn't directly improve the diversity of feature maps, nor the accuracy of the network. Through analysis, we believe that the capability of RepConv to improve the network accuracy probably benefits from the combination of batch normalization and residuals. }

\begin{figure}[h]
\centering
\subfigure[raw image]{
\begin{minipage}[t]{0.45\linewidth}
\centering
\includegraphics[width=2.85cm]{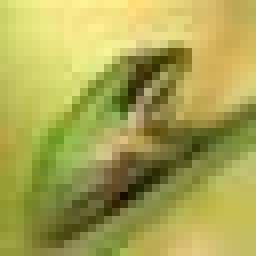}
\end{minipage}
}
\subfigure[feature map after RepConv]{
\begin{minipage}[t]{0.45\linewidth}
\centering
\includegraphics[width=5.7cm]{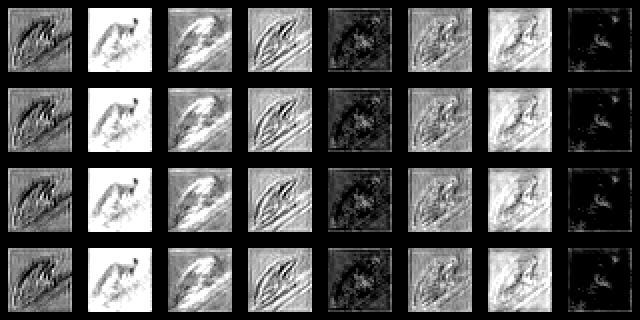}
\end{minipage}
}
\quad
\subfigure[feature map after batch normalization]{
\begin{minipage}[t]{0.45\linewidth}
\centering
\includegraphics[width=5.7cm]{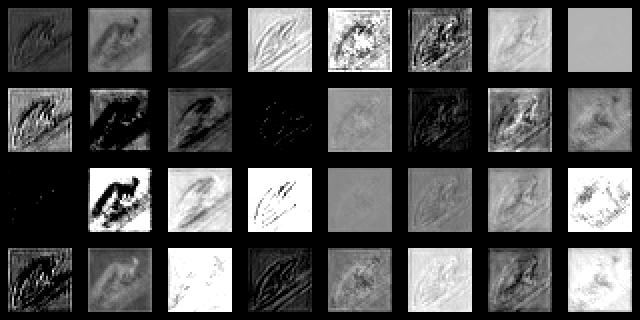}
\end{minipage}
}
\subfigure[feature map after residuals]{
\begin{minipage}[t]{0.45\linewidth}
\centering
\includegraphics[width=5.7cm]{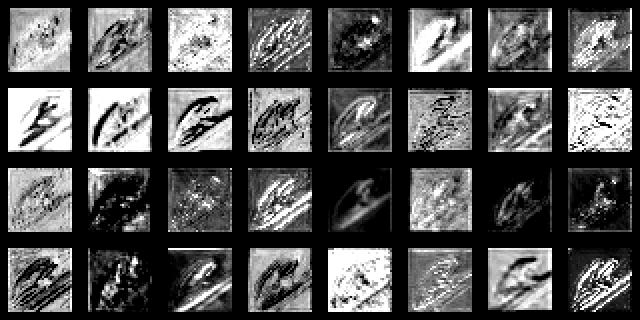}
\end{minipage}
}
\caption{Visualization of the feature map in the $7_{th}$ layer of Rep-IRNet-ResNet-20 ($\beta$=2). Given the raw input image in (a), the $7_{th}$ layer creates an feature map of 8 channels like a row of figures in (b), then RepBconv repeats the feature map $\beta^2$ times to obtain 4 rows of the same figures as (b). After the operation of batch normalization, these features are scaled and biased differently as(c), and with the help of residual link from the output of the last layer, the variance between features is even more obvious. }
\label{fig:Fig.7}
\end{figure}

To better understand of effect of batch normalization and residual link to RepBNNs, we visualize the feature map in the $7_{th}$ layer of Rep-IRNet-ResNet-20 with $\beta$=2. As displayed in Fig.\ref{fig:Fig.7}, given the raw input image in (a), the $7_{th}$ layer creates an feature map of 8 channels as a row of figures in (b), then RepBconv repeats the feature map $\beta^2$ times to obtain 4 rows of the same feature
contents as (b). After the operation of batch normalization, these features are scaled and biased differently as(c), and with the help of residual link from the output of the last layer, the variance between features is even more enlarged. Residual links accumulate the differences after batch normalization layer by layer, so that the variances between the repeated channels are continuously enlarged, which greatly increases the diversity of the feature map in Fig.\ref{fig:Fig.7}d. In this case, an informative diverse feature map of 32 channels finally contributes to a better performance of RepBNN than its originals with featur maps of 16 channels.

\hide{The output feature maps of the $7_{th}$ layer before repeat in Rep-IRNet-ResNet-20 ($\beta$=2) contains $8$ channels. RepConv repeats it for $\beta^{2}$=4 times then gets a data block containing 4 exact the same rows of feature maps, as shown in Fig.\ref{fig:Fig.7}b. After batch normalization, these data becomes obviously different, as in Fig.\ref{fig:Fig.7}c. But batch normalization just linearly transforms each channel by scaling and biasing, therefore the feature maps are still linearly related. If binarization is performed directly through Sign at this time, the data after the batch normalization just biases the threshold for binarization, the movement of which does not enrich the diversity of the feature map greatly. Wherever the residuals accumulate the differences after batch normalization layer by layer, so that the variances between the repeated channels are continuously enlarged, which greatly increases the diversity of the feature map in Fig.\ref{fig:Fig.7}d.}

\subsection{RepConv in Full-precision Network}
\label{section:full-precision}

 RepConv can be applied to full-precision convolution too. We replace the convolutions with RepConv in the original full-precision ResNet-20\cite{he2016resnet}\cite{Idelbayev18a} and a modified version with Bi-Real\cite{liu2018birealnet} structure.  In the version of Bi-Real structure, there is a residual link for each convolutional layer. However the residual links every two convolutional layers in ResNet-20.
 The test on CIFAR-10 reveals that RepConv improves the accuracy of the modified version using Bi-Real structure but reduces the performance of full-precision ResNet-20 as in Table.\ref{tab:Tab.6} . This illustrates the potential of RepConv structure in non-binary networks and its superior performance highly depends on dense residual links. 

\begin{table}[]
\caption{Benchmark RepConv ($\beta=2$) on full-precision ResNet-20 and its modified version with Bi-Real structure@ CIFAR-10 in terms of accuracy.}
\centering
\begin{tabular}{l|l|l|l}
\hline
                   & Raw(\%) & RepConv(\%) & $\Delta(\%)$                 \\ \hline
ResNet-20          & 91.73   & 90.62       & {\color[HTML]{9A0000} -1.11} \\ \hline
ResNet-20(Bi-Real) & 90.57   & 90.84       & {\color[HTML]{009901} 0.27}  \\ \hline
\end{tabular}
\label{tab:Tab.6}
\end{table}

\subsection{Comparison with SOTA BNN Methods}
\label{section:comparison}

To verify the advantages of RepBNN, we apply RepTran to a large number of recent outstanding binary neural networks with open-sourced codes. They are IR-Net\cite{qin2020irnet}, RBNN\cite{lin2020rbnn}, FracBNN\cite{zhang2021fracbnn}, and ReCU\cite{xu2021recu} with codes on CIFAR-10, and Bi-Real Net\cite{liu2018birealnet}, ReCU\cite{xu2021recu}, ReActNet\cite{liu2020reactnet}, and AdamBNN\cite{liu2021adambnn} with codes on ImageNet. We do extensive experiments on these two datasets.

\subsubsection{CIFAR-10}
The accuracy results of RepTran transformed version and the original BNNs are shown in Table.\ref{tab:Tab.7} .

\setlength{\tabcolsep}{4pt}
\begin{table}[]
\caption{The accuracy of RepBNN on CIFAR-10.($\beta=2$, Bit-width (W/A)=1/1)}
\centering
\begin{tabular}{l|l|l|l|l}
\hline
Network                     & Method        & Top-1(Raw)(\%) & Top-1(RepBNN(Ours))(\%) & $\Delta(\%)$                 \\ \hline
                            & RBNN          & 91.3           & 92.3                   & {\color[HTML]{009901} 1.0}  \\
                            & IR-Net        & 90.4           & 86.9                   & {\color[HTML]{9A0000} -3.5} \\
\multirow{-3}{*}{VGG-small} & ReCU          & 92.2           & 92.4                   & {\color[HTML]{009901} 0.2}  \\ \hline
                            & IR-Net        & 86.5           & 87.6                    & {\color[HTML]{009901} 1.1}     \\
                            & RBNN          & 86.5           & 88.6                   & {\color[HTML]{009901} 2.1}  \\
                            & FracBNN(1bit) & 87.2           & 87.6                   & {\color[HTML]{009901} 0.4}  \\
                            & RBNN-bireal   & 87.8           & 88.7                   & {\color[HTML]{009901} 0.9}  \\
\multirow{-5}{*}{ResNet-20} & ReCU          & 87.5           & 89.0                   & {\color[HTML]{009901} 1.5}  \\ \hline
                            & RBNN          & 92.2           & 93.1                   & {\color[HTML]{009901} 0.9}  \\
\multirow{-2}{*}{ResNet-18} & ReCU          & 92.8           & 93.6                   & {\color[HTML]{009901} 0.8} \\ \hline
\end{tabular}
\label{tab:Tab.7}
\end{table}
\setlength{\tabcolsep}{1.4pt}

After applying RepTran, the accuracy of ResNet-type binary networks is significantly improved. Among them, Rep-ReCU-ResNet-20 achieves an accuracy of 88.97\%, far exceeding the current state-of-the-art. 

\subsubsection{ImageNet}

We apply RepTran to Bi-Real Net\cite{liu2018birealnet}, ReCU\cite{xu2021recu}, ReActNet\cite{liu2020reactnet}, and AdamBNN\cite{liu2021adambnn}, and compare the accuracy with other popular BNN works\cite{lin2020rbnn}\cite{qin2020irnet} on ImageNet. The results are shown in Table.\ref{tab:Tab.8}. RepTran steadily improve the accuracy of the model, among which the Top-1 accuracy of Rep-AdamBNN-ReActNet-A on ImageNet has reached 71.34\%, which is the current state-of-the-art.

\setlength{\tabcolsep}{4pt}
\begin{table}[]
\begin{center}
\caption{The accuracy of RepBNN on ImageNet.($\beta=2$, Bit-width (W/A)=1/1)}
\begin{tabular}{l|l|l|l}
\hline
Network                     & Method                         & Top-1(\%)       & Top-5(\%)       \\ \hline
\multirow{8}{*}{ResNet-18}  & BNN\cite{courbariaux2016bnns}                            & 42.2            & 67.1            \\
                            & XNOR-Net\cite{rastegari2016xnor}                       & 51.2            & 73.2            \\
                            & Bi-Real Net\cite{liu2018birealnet}                    & 56.4            & 79.5            \\
                            & \textbf{Rep-Bi-Real Net(Ours)} & \textbf{57.1} & \textbf{79.5} \\
                            & IR-Net\cite{qin2020irnet}                         & 58.1            & 80.0              \\
                            & RBNN\cite{lin2020rbnn}                           & 59.9            & 81.9            \\
                            & ReCU\cite{xu2021recu}                           & 61.0            & 82.6            \\
                            & \textbf{Rep-ReCU(Ours)}        & \textbf{62.3}   & \textbf{83.8} \\ \hline
\multirow{4}{*}{ReActNet-A} & ReActNet\cite{liu2020reactnet}                       & 69.4            & 88.6            \\
                            & \textbf{Rep-ReActNet(Ours)}    & \textbf{70.2}       & \textbf{89.1}       \\
                            & AdamBNN\cite{liu2021adambnn}                        & 70.5            & 89.1            \\
                            & \textbf{Rep-AdamBNN(Ours)}     & \textbf{71.3}       & \textbf{89.8}       \\ \hline
\end{tabular}
\label{tab:Tab.8}
\end{center}
\end{table}
\setlength{\tabcolsep}{1.4pt}

\section{Conclusions}
Binary neural network (BNN) is an extreme quantization version of convolutional neural networks (CNNs) with all features and weights mapped to just 1-bit. Although BNN saves a lot of memory and computation demand to make CNN applicable on edge or mobile devices, BNN suffers the drop of network performance due to the reduced representation capability after binarization. In this paper, we propose a new replaceable and easy-to-use convolution module RepConv, which enhances feature maps through replicating input or output along channel dimension by $\beta$ times without extra cost on the number of parameters and convolutional computation. We also define a set of RepTran rules to use RepConv throughout BNN modules like binary convolution, fully connected layer and batch normalization. We apply RepTran to a large number of state-of-the-art binarization works, leading to a series of enhanced binary networks, named RepBNNs. These RepBNNs are validated on CIFAR-10\cite{krizhevsky2009cifar10} and ImageNet\cite{russakovsky2015imagenet}. Among them, Rep-ReCU-ResNet-20 achieves 88.97\% accuracy on CIFAR-10. Rep-AdamBNN-ReActNet-A achieves 71.34\% accuracy on ImageNet.

It is hoped that our work can bring some inspiration to the fields of lightweight network architecture design and network architecture search for binary neural networks.





\clearpage
%
%
\bibliographystyle{splncs04}
\bibliography{egbib}

\begin{thebibliography}{10}
\providecommand{\url}[1]{\texttt{#1}}
\providecommand{\urlprefix}{URL }
\providecommand{\doi}[1]{https://doi.org/#1}

\bibitem{bengio2013STE}
Bengio, Y., L{\'e}onard, N., Courville, A.: Estimating or propagating gradients
  through stochastic neurons for conditional computation. arXiv preprint
  arXiv:1308.3432  (2013)

\bibitem{bethge2020meliusnet}
Bethge, J., Bartz, C., Yang, H., Chen, Y., Meinel, C.: Meliusnet: Can binary
  neural networks achieve mobilenet-level accuracy? arXiv preprint
  arXiv:2001.05936  (2020)

\bibitem{bulat2020high}
Bulat, A., Martinez, B., Tzimiropoulos, G.: High-capacity expert binary
  networks. arXiv preprint arXiv:2010.03558  (2020)

\bibitem{courbariaux2016bnns}
Courbariaux, M., Hubara, I., Soudry, D., El-Yaniv, R., Bengio, Y.: Binarized
  neural networks: Training deep neural networks with weights and activations
  constrained to+ 1 or-1. arXiv preprint arXiv:1602.02830  (2016)

\bibitem{ding2019pruning}
Ding, X., Ding, G., Zhou, X., Guo, Y., Han, J., Liu, J.: Global sparse momentum
  sgd for pruning very deep neural networks. arXiv preprint arXiv:1909.12778
  (2019)

\bibitem{he2016resnet}
He, K., Zhang, X., Ren, S., Sun, J.: Deep residual learning for image
  recognition. In: Proceedings of the IEEE conference on computer vision and
  pattern recognition. pp. 770--778 (2016)

\bibitem{hinton2015distilling}
Hinton, G., Vinyals, O., Dean, J.: Distilling the knowledge in a neural
  network. arXiv preprint arXiv:1503.02531  (2015)

\bibitem{howard2019mobilenetv3}
Howard, A., Sandler, M., Chu, G., Chen, L.C., Chen, B., Tan, M., Wang, W., Zhu,
  Y., Pang, R., Vasudevan, V., et~al.: Searching for mobilenetv3. In:
  Proceedings of the IEEE/CVF International Conference on Computer Vision. pp.
  1314--1324 (2019)

\bibitem{Idelbayev18a}
Idelbayev, Y.: Proper {ResNet} implementation for {CIFAR10/CIFAR100} in
  {PyTorch}. \url{https://github.com/akamaster/pytorch_resnet_cifar10},
  accessed: 2022-01-xx

\bibitem{krizhevsky2009cifar10}
Krizhevsky, A., Hinton, G., et~al.: Learning multiple layers of features from
  tiny images  (2009)

\bibitem{lin2020rbnn}
Lin, M., Ji, R., Xu, Z., Zhang, B., Wang, Y., Wu, Y., Huang, F., Lin, C.W.:
  Rotated binary neural network. arXiv preprint arXiv:2009.13055  (2020)

\bibitem{liu2018darts}
Liu, H., Simonyan, K., Yang, Y.: Darts: Differentiable architecture search.
  arXiv preprint arXiv:1806.09055  (2018)

\bibitem{liu2021adambnn}
Liu, Z., Shen, Z., Li, S., Helwegen, K., Huang, D., Cheng, K.T.: How do adam
  and training strategies help bnns optimization? arXiv preprint
  arXiv:2106.11309  (2021)

\bibitem{liu2020reactnet}
Liu, Z., Shen, Z., Savvides, M., Cheng, K.T.: Reactnet: Towards precise binary
  neural network with generalized activation functions. In: European Conference
  on Computer Vision. pp. 143--159. Springer (2020)

\bibitem{liu2018birealnet}
Liu, Z., Wu, B., Luo, W., Yang, X., Liu, W., Cheng, K.T.: Bi-real net:
  Enhancing the performance of 1-bit cnns with improved representational
  capability and advanced training algorithm. In: Proceedings of the European
  conference on computer vision (ECCV). pp. 722--737 (2018)

\bibitem{liu2017pruning}
Liu, Z., Li, J., Shen, Z., Huang, G., Yan, S., Zhang, C.: Learning efficient
  convolutional networks through network slimming. In: Proceedings of the IEEE
  international conference on computer vision. pp. 2736--2744 (2017)

\bibitem{martinez2020real-to-binary}
Martinez, B., Yang, J., Bulat, A., Tzimiropoulos, G.: Training binary neural
  networks with real-to-binary convolutions. arXiv preprint arXiv:2003.11535
  (2020)

\bibitem{qin2020irnet}
Qin, H., Gong, R., Liu, X., Shen, M., Wei, Z., Yu, F., Song, J.: Forward and
  backward information retention for accurate binary neural networks. In:
  Proceedings of the IEEE/CVF Conference on Computer Vision and Pattern
  Recognition. pp. 2250--2259 (2020)

\bibitem{rastegari2016xnor}
Rastegari, M., Ordonez, V., Redmon, J., Farhadi, A.: Xnor-net: Imagenet
  classification using binary convolutional neural networks. In: European
  conference on computer vision. pp. 525--542. Springer (2016)

\bibitem{russakovsky2015imagenet}
Russakovsky, O., Deng, J., Su, H., Krause, J., Satheesh, S., Ma, S., Huang, Z.,
  Karpathy, A., Khosla, A., Bernstein, M., et~al.: Imagenet large scale visual
  recognition challenge. International journal of computer vision
  \textbf{115}(3),  211--252 (2015)

\bibitem{sandler2018mobilenetv2}
Sandler, M., Howard, A., Zhu, M., Zhmoginov, A., Chen, L.C.: Mobilenetv2:
  Inverted residuals and linear bottlenecks. In: Proceedings of the IEEE
  conference on computer vision and pattern recognition. pp. 4510--4520 (2018)

\bibitem{xu2021recu}
Xu, Z., Lin, M., Liu, J., Chen, J., Shao, L., Gao, Y., Tian, Y., Ji, R.: Recu:
  Reviving the dead weights in binary neural networks. arXiv preprint
  arXiv:2103.12369  (2021)

\bibitem{yang2020searching}
Yang, Z., Wang, Y., Han, K., Xu, C., Xu, C., Tao, D., Xu, C.: Searching for
  low-bit weights in quantized neural networks. arXiv preprint arXiv:2009.08695
   (2020)

\bibitem{yuan2021review}
Yuan, C., Agaian, S.S.: A comprehensive review of binary neural network. arXiv
  preprint arXiv:2110.06804  (2021)

\bibitem{zhang2018shufflenet}
Zhang, X., Zhou, X., Lin, M., Sun, J.: Shufflenet: An extremely efficient
  convolutional neural network for mobile devices. In: Proceedings of the IEEE
  conference on computer vision and pattern recognition. pp. 6848--6856 (2018)

\bibitem{zhang2021fracbnn}
Zhang, Y., Pan, J., Liu, X., Chen, H., Chen, D., Zhang, Z.: Fracbnn: Accurate
  and fpga-efficient binary neural networks with fractional activations. In:
  The 2021 ACM/SIGDA International Symposium on Field-Programmable Gate Arrays.
  pp. 171--182 (2021)

\bibitem{zhou2016dorefa}
Zhou, S., Wu, Y., Ni, Z., Zhou, X., Wen, H., Zou, Y.: Dorefa-net: Training low
  bitwidth convolutional neural networks with low bitwidth gradients. arXiv
  preprint arXiv:1606.06160  (2016)

\bibitem{zhu2019benn}
Zhu, S., Dong, X., Su, H.: Binary ensemble neural network: More bits per
  network or more networks per bit? In: Proceedings of the IEEE/CVF Conference
  on Computer Vision and Pattern Recognition. pp. 4923--4932 (2019)

\bibitem{zhuang2019Group-Net}
Zhuang, B., Shen, C., Tan, M., Liu, L., Reid, I.: Structured binary neural
  networks for accurate image classification and semantic segmentation. In:
  Proceedings of the IEEE/CVF Conference on Computer Vision and Pattern
  Recognition. pp. 413--422 (2019)

\end{thebibliography}

\end{document}